Chapter 3

**Discourse and conversation impairments in patients with dementia**

Charalambos Themistocleous

e-mail: charalampos.themistokleous@isp.uio.no

**Abstract:**

Neurodegeneration characterizes individuals with different dementia subtypes (e.g., individuals with Alzheimer's Disease, Primary Progressive Aphasia, and Parkinson's Disease), leading to progressive decline in cognitive, linguistic, and social functioning. Speech and language impairments are early symptoms in individuals with focal forms of neurodegenerative conditions, coupled with deficits in cognitive, social, and behavioral domains. This paper reviews the findings on language and communication deficits and identifies the effects of dementia on the production and perception of discourse. It discusses findings concerning (i) language function, cognitive representation, and impairment, (ii) communicative competence, emotions, empathy, and theory-of-mind, and (iii) speech-in-interaction. It argues that clinical discourse analysis can provide a comprehensive assessment of language and communication skills in individuals, which complements the existing neurolinguistic evaluation for (differential) diagnosis, prognosis, and treatment efficacy evaluation.

**Keywords:**

- clinical discourse analysis
- dementia
- discourse
- communicative competence
- emotions
- empathy
- theory-of-mind

**Preview of what is currently known:**

Neurodegeneration characterizes individuals with different dementia subtypes, leading to progressive decline in cognitive, linguistic, and social functioning. Language and communication deficits manifest early in the development of dementia (e.g., Alzheimer's Disease, Primary Progressive Aphasia, and Parkinson's Disease) affecting the production and understanding of discourse microstructure (e.g., in grammar, semantics, and pragmatics) and macrostructure (e.g., discourse planning, organizing, and structuring). This work discusses findings on discourse impairments and suggests that Clinical Discourse Analysis can provide a comprehensive assessment of language and communication skills in individuals with dementia that complements existing neurocognitive assessments for (differential) diagnosis, prognosis, and treatment efficacy evaluation.

**Objectives:**

a)  to describe the effects of dementia on discourse

b)  to identify the language and communication biomarkers for dementia assessment, diagnosis, prognosis, and treatment efficacy valuation elicited through Clinical Discourse Analysis

c)  to determine the impact of dementia on the cognitive representation of *grammar*, *communicative competence, emotions, empathy,* and *theory-of-mind*

d)  To determine whether individuals employ a socially appropriate language communication and follow the turn-taking dynamics and conventions in conversations

## 3.1  Introduction

Every year more than ten million individuals develop dementia, with almost fifty-five million people worldwide now living with dementia (1). Dementia is the progressive deterioration of cognitive, linguistic, and social functioning that affects the quality of life, including the physical, social, and economic conditions of individuals, their families, and society (2-5). Although there is no treatment for dementia, early-stage identification, and assessment of individuals with dementia are of utmost importance to enable interventions that can delay the progression of dementia and support family planning. The neurocognitive assessment aims to evaluate individuals' condition and provide early diagnosis, prognosis, and quantify intervention efficacy.

Speech, language, and communication impairments are early symptoms in individuals with dementia (6-8). For example, earlier studies have shown that discourse narratives in the autobiographies of Catholic sisters of the School Sisters of Notre Dame congregation can be an exceedingly early predictor of dementia (9). In addition, studies of the speeches of the US president Ronald Reagan (10, 11) and the comparative analysis of the British novelists Iris Murdoch and Agatha Christie works showed that narratives could provide an early prognosis of dementia development (12).

*Clinical Discourse Analysis* (CDA) examines speech, language, and communication impairments in individuals with dementia and elicits language and communication measures. These measures can provide an early, stressless, and comprehensive assessment of individuals' language and neurocognitive functioning (e.g., memory, attention, social interaction) and inform treatment approaches (13, 14). CDA involves the characterization of texts produced by individuals through language, cooperation, and social interaction in communicative settings such as conversations, semi-structured interviews (15-17), role-plays, and monologues (18).[1]

In this review, we provide evidence from recent neurolinguistic and computational developments and demonstrate that discourse provides early linguistic biomarkers for (differential) diagnosis and prognosis of individuals with dementia. We discuss the following groups of individuals:

---

[1] A more broad scope of CDA, yet uncommon in clinical settings, is the study of meaningful symbolic behavior of individuals in any mode, including social structures expressed through discourse, e.g., the discourse of race and power 19. Blommaert J. Discourse. Cambridge: Cambridge University Press; 2005.

i. individuals with Primary Progressive Aphasia (PPA), a progressive neurological condition, which primarily affects speech and language. Individuals with PPA are grouped into three variants based on their distinct underlying neuropathology and area of brain damage (20). According to current classification criteria, their characteristic neuropathology and damage patterns give rise to different discourse deficits across three variants (21, 22), namely in individuals with the non-fluent PPA variant (nfvPPA), individuals with the semantic PPA variant (svPPA), and individuals with the logopenic PPA variant (lvPPA);

ii. individuals with Alzheimer's Disease (AD) constitute the larger group of individuals with dementia. They are characterized by a progressive deterioration of memory, language, conversation, and ability to perform everyday activities, unlike individuals with Mild Cognitive Impairment, whose cognitive impairments are incipient and retain their day-to-day functioning.

iii. individuals with Parkinson's Disease (PD) are characterized by a progressive deterioration of movement functioning, which impairs balance, speaking, language, chewing, and swallowing.

We identify the effects of brain damage due to neurodegeneration on discourse microstructure (e.g., phonology, morphology, and syntax), macrostructure (e.g., *cohesion* of linguistic forms to determine whether individuals produce a text that follows the grammar and *coherence* of meanings and whether text productions make sense (23, 24). Macrostructure and microstructure are intertwined and often the attempt to disentangle them is difficult as the same constituents can perform both microstructure and macrostructure functions, yet the distinction is necessary for the description of discourse structure. Although, most language impairments associate with left hemisphere damage (21, 25), as discourse involves speech (and writing (26, 27)), language, emotions, social cognition, and cognitive domains, such as memory and attention, discourse impairments can results on neurodegenerative effects on the left and right hemispheres (28-31).

Here, we will discuss findings concerning the following four areas: (i) *language function, cognitive representation,* and *impairment*[2], and examine how dementia impacts the cognitive representations of grammar (rules and principles) that enable speakers to produce grammatically correct sentences (32); (ii) *communicative competence, emotions, empathy,* and *theory-of-mind* (33) and evaluate whether individuals employ a socially appropriate language communication; and (iii) *talk-in-interaction* to identify how individuals with dementia follow the turn-taking dynamics and conventions in conversations (34-36).

## 3.2 Language function, cognitive representation, and impairment

### 3.2.1 *Discourse microstructure and dementia*

Individuals with dementia produce speech with deficits that affect speech and articulation (e.g., prosodic patterns and rhythmical patterns), phonology (e.g., phonological errors, such as insertions, deletions, syllable structure simplifications), prosodic phonology (stress, rhythmical errors, intonation), morphology (e.g., morphological errors in verb and noun inflections, tense, and number agreement, grammatical and content word production, and parts of speech) (37), lexicon (38), syntax (e.g., phrase structure and embedded phrases) (16, 39, 40), and semantics (e.g., lexical semantics, naming) (6, 41-48). These deficits can characterize individuals with other speech and language disorders as well, such as stroke aphasia (49-52). Discourse analysis employs both manual analysis and computational methods, such as acoustic analysis, natural language processing, and machine learning to automate the analysis (20, 53-60).

---

[2] Researchers often use the word "errors" to refer to incorrect productions with respect to correct productions (targets) in typical speakers; however, individuals produce language that follows their own grammatical system or interlanguage as it has developed after the brain damage. Moreover, the term error conceals the systematicity of these productions.

Traditionally, these discourse microstructure impairments are assessed during conventional neurophysiological examination with standardized language assessment tasks and neurolinguistic batteries, such as the Boston Naming Test (BNT; 61), Western Aphasia Battery-Revised (WAB-R; Kertesz (62)), Boston Diagnostic Aphasia Examination (BDAE; 63), Psycholinguistic Assessment of Language Processing in Aphasia (PALPA; 64); the Verb and Sentence Test (VAST; 65). Nevertheless, single-domain standardized tests (cf., articulation, and conformation naming) are not meant to assess broad language communication skills (e.g., articulation, morphosyntax, semantics, pragmatics, turn-taking), which alternatively require the application of multiple separate time-consuming and stressful language assessment tests. Thus, CDA aims to provide a comprehensive analysis and assessment of speech and grammar in context without requiring lengthy evaluation tests to target specific language domains. Here, we will review the primary microstructural deficits in individuals with PPA, AD, and PD, using combined information from studies employing discourse and standardized test evaluations.

*In individuals with PPA,* the primary language deficits correspond to the effects of neurodegeneration on the left hemisphere (43, 66-70). Specifically, neuroimaging data shows that the peak atrophy site in individuals with nfvPPA is the posterior inferior frontal gyrus (pIFG), also known as Broca's area. In individuals with svPPA, the peak atrophy is the left anterior temporal lobe, and in individuals with lvPPA, the peak atrophy is located in the left posterior temporal and inferior parietal regions (21, 44, 71, 72).

More specifically, the individuals with nfvPPA, svPPA, and lvPPA differ in their discourse microstructure deficits. First, the individuals with nfvPPA are characterized by agrammatism resulting in telegraphic speech productions, namely they omit grammatical words, such as conjunctions, particles, and prepositions. Often their speech is accompanied by Apraxia of Speech (AOS), which is associated with slow effortful speech with speech errors and pauses (73-77). Individuals with nfvPPA with agrammatism are characterized by substantial deficits in function word production (17, 78, 79). Consequently, in the context of producing discourse, individuals with nfvPPA produce more filled pauses than individuals with the semantic variant as they strive to construct grammatical structures and words (56). Studies of the connected speech productions showed impaired production of sentence structure components, such as verb and noun phrases (7, 40). In addition, individuals with the nfvPPA are characterized by syntactic comprehension impairments, especially during the perception of syntactically complex sentences (80). Supporting evidence from naming tests showed that single-word comprehension and object naming are retained (21).

Although, AOS and agrammatism are the two key diagnostic features in individuals with nfvPPA, agrammatism often occurs without AOS (78) and AOS without agrammatism (81). Consequently, many studies distinguish individuals with the agrammatic variant of PPA (78) and individuals with PPA and AOS (PPAOS) (82-84). Often the classification is unclear as symptoms progress, leading to different degrees of language deterioration (85).

Moreover, individuals with nfvPPA produce selectively fewer verbs than healthy individuals (17, 78, 79) but verb perception can be preserved (86). Verb production may be preserved in individuals with other PPA variants, although individuals with lvPPA show deficits during discourse in noun production (17, 78, 79). In a computational study of morphology in individuals with PPA, Themistocleous et al. (37) have shown differential usage of parts of speech in individuals with different PPA variants (20).

Second, individuals with svPPA are characterized by impaired confrontation naming and single-word comprehension, impaired object knowledge, dysgraphia, and dyslexia (37, 43, 87). However, unlike individuals with nfvPPA, speech production is spared in individuals with svPPA. Also, individuals with svPPA are characterized by deficits in inflectional morphology. For example, Wilson et al. (88) showed that individuals with svPPA are impaired in inflecting low-frequency irregular words (89). Nonetheless, inflectional morphology can be impaired in individuals with the other two PPA variants as individuals with nfvPPA show difficulties in inflecting pseudowords and individuals with lvPPA display morphophonological deficits (88).

Third, individuals with lvPPA are characterized by impaired single-word retrieval in spontaneous speech and naming and impaired sentence and phrase repetition, often with phonological errors. Further analysis of discourse has the potential to reveal interactions between lexical and morphosyntactic categories as suggested by task-based assessments (90).

*Individuals with AD and MCI* are impaired in discourse microstructure; although, their deficits are incipient in individuals with MCI and become progressively more severe in individuals with AD, which may end into mutism—although, mutism is more common in individuals with frontotemporal dementia (91, 92).

Studies using signal processing have shown that individuals with AD and MCI produce connected speech productions with significantly different patterns in segmental acoustic structure (i.e., vowels and consonants); prosody, voice quality, speech fluency and speech rate and demonstrated that speech acoustics can be employed both for diagnosis and differential diagnosis of individuals with AD and MCI from healthy controls and provide classification models for diagnosis or subtyping (54, 93-96). For example, a recent study by Themistocleous et al. (57) found significantly slower speech productions in Swedish individuals with MCI than healthy controls, manifested as slower speech rate and long syllables. Moreover, they found that the speech of individuals with MCI is characterized by a greater degree of breathy voice, dysphonia, center of gravity, and shimmer than in healthy controls. They argue that the acoustic differences of individuals with MCI from healthy individuals indicate a physiological impairment in the fine control of vocal fold vibration, pulmonary pressure, respiration, and the coordination of phonation and articulation [15, 50–53]. However, other studies show mixed results. More specifically, studies on emotional prosody showed impaired prosody in expressive speech productions, such as less pitch modulation and slower speech rate in individuals with dementia than in healthy controls but their ability to control pitch and speech rate was normal (97).

Although phonology is relatively intact in individuals with MCI and AD, several studies showed phonetic and phonological errors, such as incorrect phoneme production, false starts, phonological paraphasias, and articulatory difficulties (98, 99). Compared to healthy controls, individuals with AD can exhibit deterioration or simplification of grammar and semantics (100).

For example, an early study of narratives and constrained tasks showed significant errors in open and closed lexical classes, pronouns and morphosyntax (e.g., inflection and agreement) between individuals with AD and healthy elderly individuals (101). Similarly, a study with Greek individuals showed that individuals with probable AD, were more impaired in verb aspect than in tense and agreement compared to healthy controls, in both production and grammaticality judgement; in contrast, verb agreement was in general retained (102). Furthermore, individuals with AD can produce discourse with word finding and lexical retrieval difficulties (103), redundant words, and a higher proportion of closed-class words (98, 104).

*Individuals with PD* are characterized by speech acoustic differences in speech production and intonation identified both from discourse and non-discourse data (105, 106) and in other linguistic domains, such as syntax and sentence production (107). Moreover, individuals with the behavioral variant of frontotemporal dementia (bvFTD) manifest progressive changes in personality, behavior, and social cognitive functions (108, 109) but can also manifest language impairments in their lexicon, semantics, prosody, reading, and writing (110). At the same time, they may preserve motor speech production and morphosyntax (110).

### *3.2.2 Discourse macrostructure and dementia*

In conversations, individuals with dementia, especially those with PPA, display impairments in discourse planning and macrostructure. Glosser and Deser (111) suggested that individuals with dementia are selectively more impaired in discourse macrostructure (e.g., thematic coherence and cohesion) than microstructure (e.g., phonology, morphology, lexicon, and syntax). Moreover, as dementia develops, text cohesion and coherence (112) become progressively more impaired (113).

*Cohesion* impairments manifest as irregularities in how individuals establish cohesive relationships in the text (114). For example, individuals with dementia produce text with impaired lexical cohesion (i.e., lexical repetition and lexical chains, collocation), discourse markers, which connect post-sentential constituents, having additive (e.g., and, furthermore, in addition), adversative (e.g., but, however, nevertheless), causal (e.g., so, consequently), and temporal meaning (e.g., then, after that, finally). Moreover, they display compromised application of cohesive devices, such as anaphora and cataphora (115), namely referencing usually with pronouns to an earlier (anaphora) or subsequent (cataphora) name or entity in discourse), substitution, conjunction, and replacement (116).

Furthermore, individuals with dementia display deficits in making cohesive associations with adjacency pairs, such as question-answer pairs, enumerations, greeting-greeting pairs, invitation-acceptance or rejection, and request-acceptance or rejection. When speakers employ adjacency pairs, the first part of the adjacency pair creates expectations that an ensuing part should satisfy, e.g., in a question-answering pair, listeners expect an answer to a question and in an enumeration an utterance, such as "I am going to state three things" should be followed by a list of three things; missing these associations are common in patients with dementia. For example, Ramanathan (117) notes that the expectations developed by Tina, an individual with AD are not fulfilled by her, and the communication collapses:

> *"Tina' s talk here does not allow stanza parsing. Her talk starts off as a narrative (…) where she talks about how her India trip came about ("I had always wanted to go to India ..."), but she does not sustain her effort. In fact,*

> *in some instances she does not respond at all. (…) she does not pick up my prompts as cues for her to keep talking and (…) she is non-responsive to my question (…) but, once again, she does not develop her utterances into narratives"* (103).

*Coherence* is the structuring and continuity of meanings and the semantic relationship of oral (or written) productions to their context including the situational conditions related to space, time, participants, and sociocultural meanings (118). Coherence deficits reveal impairments in memory and the semantic-linguistic interface, including impairments in recalling and organizing semantic meanings, such as past experiences and knowledge about the world (119). Individuals with AD and MCI produce discourse with impaired semantic meaning and structure (119).

In addition, individuals with svPPA manifest semantic impairments in discourse affecting production of lexical items and particularly content words (120). A recent study by Seixas Lima et al. (121) on individuals with svPPA showed that although they produced episodic information related to the discourse topic, the semantic information was unrelated to discourse topic. The authors argue that, for individuals with svPPA, impairment depends on the selection of relevant semantic information and the inhibition of irrelevant ones. These findings are consistent with evidence from confrontational naming tasks, such as the Boston Naming Test (BNT) (61) and the Hopkins Assessment on Naming Actions (HANA) (122).

Furthermore, patients with dementia produce speech with impaired information packaging, concerning the new and old information, contrast, and pragmatic implicatures (123). Specifically, information packaging is achieved using linguistic means such as syntax and prosody (124, 125). Information packaging using syntax is manifested with cleft structures, where constituents are moved to a different position in the sentence to express contrast, emphasis, etc., as in the following examples in (1):

(1)
  i. It is George[F][3] who went to the movies last week.
  ii. It was to the movies[F] that George went last week.
  iii. It was yesterday[F] that Jerry went to the movies.
  iv. What I need is a nice milk chocolate.

The cleft structures in the examples above can imply a contrasted constituent as:

(2)
  i. It was GEORGE who went to the movies last week [not Maria].

Moreover, prosody (e.g., intonation and phrasing cues) express information structure in English and other languages. For example, new-information focus and contrast are manifested using nuclear pitch accents that highlight the prominent or contrasting constituent (126-129), whereas preceding words, if any, are marked with a different type of accent, a.k.a. prenuclear pitch accents.

---

[3] [F] indicates the focused or highlighted constituent.

In English syntax expresses information structure (cf. examples above (1) and (2)) combined with a nuclear pitch accent at the end manifesting broad focus (3):

(3)
i. George went to the movies last week.

These cases demonstrate an interplay between prosody and syntax. Dementia can impair prosody and syntax in individuals with non-fluent PPA with agrammatism (124), individuals with MCI and AD (54, 95, 110, 130).

Using Rhetorical Structure Theory, Abdalla et al. (131) showed the effects of AD on discourse rhetorical mechanisms (132). RST evaluates various relationships and how they are constructed in discourse, e.g., Elaboration, Circumstance, Solutionhood, Cause, Restatement, and presentational relations, Motivation, Background, Justify, and Concession (132, 133). Studies of discourse in the Nun Study, a longitudinal study of cognitive decline, also showed that nuns with AD produced discourse with impaired idea density; in fact, research on the nun study showed that discourse deterioration is a very early predictor of dementia (134).

Overall, discourse macrostructure impairments can correlate with the deterioration of language and cognition, such as working memory, planning, generation, problem-solving, and abstraction (111).

### 3.3 Dementia and communicative competence, emotions, empathy, and theory-of-mind

Communication assessment aims to determine how individuals employ language in the appropriate social context, follow social norms, and make connections with participants and settings. Therefore, Hymes (135) suggests utterances should be evaluated concerning their discourse context. In other words, the communicative competence of individuals is assessed based on whether they produce utterances that consider the following discourse settings, namely:

1. the other participants;
2. the roles they assume;
3. the conversational topic;
4. the communicative channel (e.g., writing and speaking);
5. the language code (e.g., language variety, dialect, language style);
6. the message form (e.g., lecture, a conversation, a fairy tale, narrative);
7. the situation (e.g., a ceremony, a friendly conversation);
8. the purpose of the speech; and
9. the key (e.g., tone and manner).

Therefore, there is a fundamental distinction between CDA and standardized language assessment in that CDA assesses language communication, which involves several distinct components at once, such as the following: (i) the *intentionality of discourse* and whether individuals perceive utterances as intentional and actively make a cooperative effort to produce and understand the discourse content (114, 136); (ii) *the situationality of discourse* and whether individuals produce utterances that are related to the immediate discourse context, and (iii) the *intertextuality* of discourse and whether individuals' utterances connect discourse productions to the broader intertextual context correctly (114).

Kong et al. (137) employed story grammars, which consider information such as the conversational background, participants, and the time, and place of conversation to analyze discourse produced by individuals with fluent aphasia, non-fluent aphasia, AD, and healthy controls. Their study showed significant differences in the production of situational discourse information in individuals with AD and healthy controls. Interestingly, their study showed similarities in the use of situational information between individuals with AD and individuals with fluent aphasia.

Individuals with dementia can exhibit discourse impairments in communicative competence (138). However, individuals with the behavioral variant of frontotemporal dementia (bvFTD) show communicative competence impairments more predominantly than other individuals with dementia. These individuals display impairments in early behavioral disinhibition, which is the distinguishing symptom of individuals with bvFTD clinical syndrome from individuals with AD, dementia with Lewy bodies, and vascular dementia (108). Inappropriate language accompanies an overall loss of manners of decorum, such as cursing, speaking loudly, express offensive, sexually, obscene remarks, jokes, and opinions (108). In addition, apathy, failure to initiate or sustain a conversation, loss of empathy, insight, and executive dysfunction are associated with frontal and temporal atrophy (108). Also the speech of individuals with bvFTD is characterized by selective impairments linked to an overall degraded communicative competence, as reflected by poor organization discourse, simplification of grammatical production, selective impairments in word use, and changes in acoustic properties (110, 138-142).

Notably, the assessment of communicative competence brings language to the fore as the connecting link between the social context, paralinguistic expression and emotion, empathy/sympathy, and theory of mind (ToM). ToM is the ability of individuals to attribute mental states to other individuals and employ the states to understand and predict actions and discourse contributions. Using standardized tests for memory, comprehension, and general inferencing question, Youmans and Bourgeois (143) found that individuals with mild to moderate AD exhibit mild but specific ToM impairment. Again, ToM impairments are more severe in individuals with bvFTD. However, standardized neurocognitive testing of ToM can only modestly distinguish between individuals with FTD and AD (144), as these tests create an artificial environment that does not provide the necessary social context to adequately evaluate ToM. In contrast, discourse can provide a natural environment and comprehensive assessment of ToM.

These findings from discourse provide quantitative measures of communicative competence and assess social and behavioral symptoms using more naturalistic and conversational interactions than standardized language tests (e.g., identifying pictures in cards or picture books and repeating words and sentences).

## 3.4 Dementia and talk-in-interaction (turn-taking)

As discourse is used in conversations between individuals and other individuals (e.g., clinicians and other individuals), conversation analysis can be employed as a method to identify how dementia influences both language and social interaction. Conversation analysis studies the social conventions that facilitate the interaction of interlocutors and the passing of conversational turns from one participant to another. The conversational turns are the basic units of any conversation (36). The aim of conversation analysis in the clinic is twofold: first, it aims to determine whether the basic properties that characterize social interaction break down in individuals with dementia and quantify the communicative characteristics of their speech and second, it aims to identify how individuals with dementia construct the conversational turn vis-à-vis other groups of speakers (e.g., healthy controls, individuals with different conditions or with respect to an earlier stage of the same individual). Examining how individuals with dementia engage in conversations in the clinic provides information about the language and social interaction impairments.

First formulated by Sacks et al. (36), conversation analysis aims to determine whether speakers follow the social conventions that regulate the exchange of turns in conversations. In particular, the authors formulated a set of simple principles that determine conversation (36): speaker change recurs; one party speaks at a time; occurrences, when speakers talk simultaneously, are common but brief; turn transitions with a slight or no gap make the majority of turn transitions; the turn order and size are not fixed, and the length and content of transitions vary. Moreover, the selection of speakers in the conversation follows conventionalized turn-allocation techniques, which are a core component of any conversation. The conversation exchange consists of turns structured from turn-constructional units (cf. sentences consisting of phrases and words). A conversational unit contains turn transition points where the current speaker can select the next speaker or decide to continue, known as self-selection, using gestures such as prosody or intonation, hand and head gestures, body posture, glance, and eyebrow movement. Lexicogrammar can also indicate transitions, such as specific lexical units or the right end of a sentence. For example, conversation exchange involves the identification of the appropriate places to exchange the conversational turn, such as passing it to the next speaker and resolving conflicts that may occur.

Individuals with PPA display interactional difficulties during communication. For example, a study of individuals with svPPA showed that although these individuals participated actively in the conversation, they had problems maintaining the flow of interaction, such as requesting for confirmation and displayed an inability to keep with the conversation, such as to initiate or continue the conversational topic (145). In addition, individuals with AD differ in the construction of turns from healthy controls. For example, a study of conversations in individuals with AD showed that individuals with AD produced fewer word per turn and fewer speech acts—especially, requestives, assertives—than healthy individuals (146).

The types of conversations, e.g., casual conversations (147), telephone conversations (148), map task navigation, computer-mediated decision-making interactions, and spontaneous dialogue data (149) determine the role of exchange between speakers. However, a common complaint is that individuals with dementia often have difficulty following conventional patterns, such as ring-greeting-message-greeting, that characterize phone conversations (150).

Moreover, reduced conversational skills characterize individuals with dementia, including PPA, MCI, AD, and PD (151-154). For example, studies showed that individuals with PPA could "maintain turn-taking but had reduced amount of talk and were able to request confirmation and actively repair their own and their partners' trouble in talk" (155). Nevertheless, studies using conversational analysis usually rely on a few participants. Thus, it is far from straightforward to generalize on the population.

Whitworth et al. (156) employed conversational analysis and showed that individuals with PD display impairments in turn initiation, turn-taking, and repair, such as failing or delaying responding to conversational cues when turns are allocated to them by the current speaker. As a result, in individuals with PD, conversational difficulties arise regarding speaker coordination during turn-taking and turn-resolution (157, 158). These failures may occur due to a failure to perceive a turn-taking cue during the interaction. Several studies in talk-in-interaction benefit from analyzing multimodal cues from speech language and video. However, these studies generally rely on very few participants (159); thus, it is essential to employ conversation multimodal analytic research on larger individual groups to safeguard the generalizability of the findings.

## 3.5  Conclusions

This paper has discussed discourse and conversation in the context of assessment and diagnosis and demonstrated that CDA provides information about speech communication impairments across discourse domains. The multimodal information highlights the value of CDA as an approach that provides measures that can complement those from current standardized language evaluation batteries (160-164) for assessment, diagnosis, prognosis, and therapy efficacy estimation (159, 163-165).

## 3.6     Major takeaways

i.  Dementia affects speech, language, and communication in most individuals with dementia, but these are especially evident in individuals with Primary Progressive Aphasia.
ii. Deficits characterize the microstructure (e.g., phonology, morphology, syntax) and discourse macrostructure (e.g., cohesion, coherence), theory of mind, and conversation.
iii. Discourse analysis in individuals with dementia provides comprehensive linguistic biomarkers for speech, language, communication, and cognition assessment, (differential) diagnosis, prognosis, and treatment efficacy valuation.

# Page 19

121. Seixas Lima B, Levine B, Graham NL, Leonard C, Tang-Wai D, Black S, Rochon E. Impaired coherence for semantic but not episodic autobiographical memory in semantic variant primary progressive aphasia. Cortex. 2020;123:72-85.
122. Breining BL, Tippett DC, Davis C, Posner J, Sebastian R, Oishie K, Hillis AE. Assessing dissociations of object and action naming in acute stroke. Clinical Aphasiology Conference; Monterey, CA2015.
123. Roberts A, Savundranayagam M, Orange JB. Non-Alzheimer Dementias. In: Cummings L, editor. Research in Clinical Pragmatics. Cham: Springer International Publishing; 2017. p. 347-77.
124. Walenski M, Mack JE, Mesulam MM, Thompson CK. Thematic Integration Impairments in Primary Progressive Aphasia: Evidence From Eye-Tracking. Front Hum Neurosci. 2020;14:587594.
125. Ribu ISB. Language and cognition in healthy aging and dementia. Olslo: University of Oslo; 2019.
126. Themistocleous C. Seeking an Anchorage. Stability and Variability in Tonal Alignment of Rising Prenuclear Pitch Accents in Cypriot Greek. Language and Speech. 2016;59(4):433-61.
127. Themistocleous C. Prosody and Information Structure in Greek (Prosodia kai plirophoriaki domi stin Ellinici). Athens: National and Kapodistrian University of Cyprus; 2011.
128. Beckman M, Pierrehumbert J. Intonational Structure in Japanese and English. Phonology Yearbook. 1986;3:255-309.
129. Pierrehumbert J. Phonology and Phonetics of English intonation. Boston: Massachusetts Institute of Technology; 1980.
130. Misiewicz S, Brickman AM, Tosto G. Prosodic Impairment in Dementia: Review of the Literature. Current Alzheimer Research. 2018;15(2):157-63.
131. Abdalla M, Rudzicz F, Hirst G. Rhetorical structure and Alzheimer's disease. Aphasiology. 2018;32(1):41-60.
132. Taboada M, Mann WC. Rhetorical Structure Theory: looking back and moving ahead. Discourse Studies. 2006;8(3):423-59.
133. Kong AP-H, Linnik A, Law S-P, Shum WW-M. Measuring discourse coherence in anomic aphasia using Rhetorical Structure Theory. International Journal of Speech-Language Pathology. 2018;20(4):406-21.
134. Farias ST, Chand V, Bonnici L, Baynes K, Harvey D, Mungas D, Simon C, Reed B. Idea density measured in late life predicts subsequent cognitive trajectories: implications for the measurement of cognitive reserve. J Gerontol B Psychol Sci Soc Sci. 2012;67(6):677-86.
135. Hymes DH. Ethnography, Linguistics, Narrative Inequality Toward An Understanding Of voice Critical Perspectives on Literacy and Education. New York: Routledge; 1996.
136. Grice HP. Logic and Conversation. In: Cole P, Morgan JL, editors. Syntax And Semantics Speech Acts. 3: Academic Press; 1975. p. 41-58.
137. Kong AP, Whiteside J, Bargmann P. The Main Concept Analysis: Validation and sensitivity in differentiating discourse produced by unimpaired English speakers from individuals with aphasia and dementia of Alzheimer type. Logoped Phoniatr Vocol. 2016;41(3):129-41.
138. Rosen HJ, Allison SC, Ogar JM, Amici S, Rose K, Dronkers N, Miller BL, Gorno-Tempini ML. Behavioral features in semantic dementia vs other forms of progressive aphasias. Neurology. 2006;67(10):1752-6.
139. Carr AR, Ashla MM, Jimenez EE, Mendez MF. Screening for Emotional Expression in Frontotemporal Dementia: A Pilot Study. Behavioural Neurology. 2018:1-6.
19

**Biographical note**

*Charalambos Themistocleous*, Associate Professor of speech, language, and communication at the Department of Special Needs Education at University of Oslo. He served as a postdoctoral researcher in computational neurolinguistics at the Department of Neurology at Johns Hopkins University. He is a computational linguist with research interests in speech, language, and communication. He is interested in the relationship between the brain and language and in providing opportunities to patients with aphasia by developing automated, personalized, high-quality speech and language assessments, diagnoses, and prognoses. He has developed machine learning models for automatic treatment, monitoring, and differential diagnosis of individuals with Primary Progressive Aphasia, Alzheimer's disease, and Mild Cognitive Impairment from speech, language, structural/functional MRI, and neurophysiological testing. [ORCID: 0000-0002-9138-0939]